\documentclass{dukecei}
\usepackage{amsmath,amsfonts,amssymb,amsthm}
\usepackage{algorithm}
\usepackage{algorithmic}
\usepackage{amssymb}
\usepackage{bbm}
\usepackage{bm}
\usepackage{graphicx}
\usepackage{multirow}
\usepackage{mathtools}
\usepackage{subcaption}
\usepackage{textcomp}
\usepackage{xcolor}
\usepackage[inline]{enumitem}
\usepackage{booktabs}
\usepackage{colortbl}
\usepackage{caption}
\usepackage[table]{xcolor}
\usepackage{booktabs} 
\usepackage{longtable} 
\usepackage{tocloft} 
\usepackage{makecell, multirow} 
\usepackage{float}

 \usepackage{newtxtext,newtxmath} 
 \usepackage{geometry}
\usepackage{booktabs}
\usepackage{threeparttable}
\usepackage[table]{xcolor}
\usepackage{array}

\definecolor{prefillbg}{RGB}{243,250,245}   
\definecolor{decodebg}{RGB}{241,249,252}    
\definecolor{speedlrbg}{RGB}{243,246,253}   
\definecolor{speeddensebg}{RGB}{252,246,239}

\newcolumntype{P}{>{\columncolor{prefillbg}[\tabcolsep][\tabcolsep]}r}
\newcolumntype{D}{>{\columncolor{decodebg}[\tabcolsep][\tabcolsep]}r}
\newcolumntype{L}{>{\columncolor{speedlrbg}[\tabcolsep][\tabcolsep]}r}
\newcolumntype{E}{>{\columncolor{speeddensebg}[\tabcolsep][\tabcolsep]}r}

\usepackage{booktabs,multirow}
\usepackage{arydshln}
\usepackage[table]{xcolor}
\definecolor{hlblue}{HTML}{D9F2F2} 

\usepackage{booktabs,multirow}
\usepackage{arydshln}
\usepackage[table]{xcolor}
\definecolor{hlblue}{HTML}{D9F2F2} 

\usepackage[utf8]{inputenc}

\usepackage{booktabs}
\usepackage{threeparttable}
\usepackage{makecell}
\usepackage{siunitx}
\usepackage[table]{xcolor}
\usepackage{array}

\definecolor{RowBlue}{RGB}{235,245,255} 
\usepackage{geometry}
\usepackage{parskip}

\usepackage{times}  
\usepackage{helvet}  
\usepackage{courier}  
\usepackage{graphicx} 
\usepackage{natbib}  
\usepackage{caption} 
\usepackage{amsmath,amsfonts,amssymb,amsthm}
\usepackage{amssymb}

\usepackage{bm}
\usepackage{graphicx}
\usepackage{multirow}
\usepackage{mathtools}
\usepackage{subcaption}
\usepackage{textcomp}
\usepackage[inline]{enumitem}

\usepackage{cleveref} 
\usepackage{booktabs}
\usepackage{colortbl}
\usepackage{caption}
\usepackage[utf8]{inputenc}
\usepackage{booktabs} 
\usepackage{longtable} 
\usepackage{tocloft} 
\usepackage{makecell, multirow} 
\usepackage{algorithm}
\usepackage{algorithmic}

\usepackage{float}
\usepackage{graphicx}
\usepackage[utf8]{inputenc}
\usepackage{placeins}

\usepackage{amssymb}   

\usepackage{soul}

\title{FlashSVD v1.5: Making Low-Rank Transformers Inference Actually Fast}

\author{
    Wenhao Wu$^{1,*}$, 
    Zishan Shao$^{1,2,*,\dagger}$\footnote[0]{$^*$ Equal Contribution}\footnote[0]{$^\dagger$ Correspondence E-mail: zishan.shao@duke.edu},
    Kangning Cui$^{4}$, 
    Jinhee Kim$^{1}$, 
    Yixiao Wang$^{2,3}$, 
    Hancheng Ye$^{3}$,  
    Danyang Zhuo$^{3}$,    
    Yiran Chen$^{2}$\\[1em]
    \normalsize $^{1}$Department of Electrical \& Computer Engineering, Duke University \\
    \normalsize $^{2}$Department of Statistical Science, Duke University \\
    \normalsize $^{3}$Department of Computer Science, Duke University \\
    \normalsize $^{4}$Department of Computer Science, Wake Forest University \\
}

\begin{document}

\maketitle
\thispagestyle{firstpagestyle} 

\begin{abstract}
SVD-based Low-rank compression reduces transformer parameters and nominal FLOPs, but these savings often translate poorly into real LLM serving speedups. We show that this gap is largely a runtime problem: factorized checkpoints fragment execution paths, and the resulting overhead differs substantially between prefill and autoregressive decode. We present \textbf{FlashSVD v1.5}, a unified inference runtime for serving SVD-compressed transformers. FlashSVD v1.5 maps diverse public SVD compression families to a common factorized representation and combines phase-specific kernels with dense-KV decode, packed MLP execution, and per-layer CUDA-graph replay to reorganize the low-rank serving path into a thin runtime. Across representative decoder-serving settings, FlashSVD v1.5 achieves up to $2.55\times$ decode and $2.39\times$ end-to-end speedup, and it attains $1.48\times$ average decode and $1.44\times$ average end-to-end speedup across multiple popular SVD compression families. These results suggest that practical low-rank acceleration requires runtime co-design, not compression algorithms alone. Our code is available at: \url{https://github.com/Zishan-Shao/FlashSVD}.
\end{abstract}


\section{Introduction}

Modern Large Language Models (LLMs) achieve strong performance across diverse tasks, but their rapidly increasing size makes deployment increasingly difficult. Singular value decomposition (SVD) has recently emerged as a prominent direction for LLM compression \cite{yuan2023asvd,wang2025svdllm,gao2026dfsvd,gao2025gfsvd,hu2026saes,ding2025dipsvd,lin2024modegpt,wang2025svdllmv2,wang2025dobisvd,li2025adasvd,mi2025drank}. These approaches reduce model size and nominal FLOPs while remaining hardware-friendly. In practice, however, these theoretical savings often do not translate into comparable inference speedups.

This gap arises from a mismatch between low-rank checkpoints and serving runtimes. First, fragmented checkpoint formats, such as SVD-LLM v1/v2 \cite{wang2025svdllm,wang2025svdllmv2} and Basis Sharing \cite{wang2024basissharing}, require separate runtime handling. Second, factorizing a large weight into smaller operators increases kernel boundaries and launch overhead. Third, prefill and autoregressive decode expose different bottlenecks: prefill mainly benefits from reduced arithmetic, whereas decode is memory-bound and repeatedly pays for history-dependent low-rank computation. To address these issues, we introduce \textbf{FlashSVD v1.5}, a unified high-performance inference runtime for low-rank transformers. It supports multiple public checkpoint families under one execution framework and combines phase-specific execution with dense-KV attention, packed MLP execution, and per-layer CUDA-graph replay to reorganize the fragmented low-rank path into a thinner serving path.

Experiments on LLaMA-family decoders show that FlashSVD v1.5 improves decode and end-to-end latency on fixed low-rank checkpoints, with up to $2.55\times$ decode and $2.39\times$ end-to-end speedups for autoregressive LLMs. These gains transfer across public checkpoint families and remain visible over long decode horizons. Under practical bf16 cached decoding, FlashSVD remains close to the baseline cached decode path, although we do not claim token-by-token identity. Additional encoder results are reported in the appendix. Together, these results show that making low-rank models actually fast requires system-level co-design, not compression algorithms alone.

\begin{figure*}
    \centering
\includegraphics[width=0.95\linewidth]{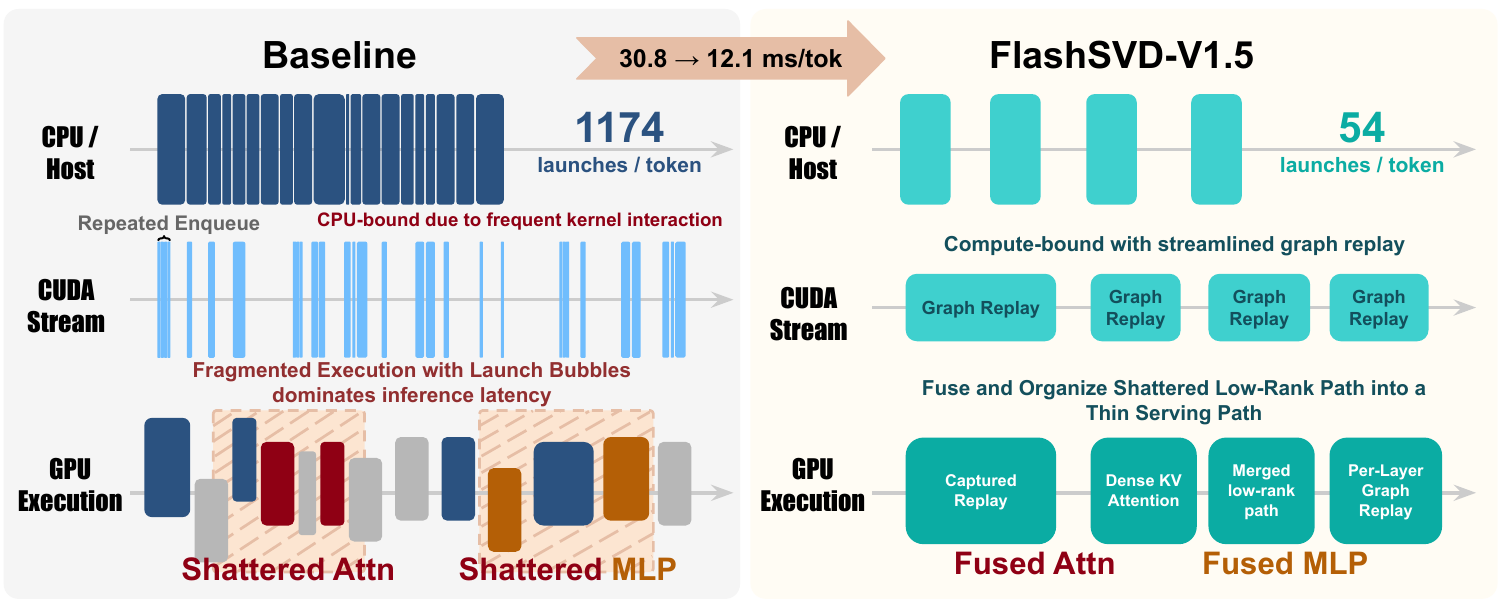}
    \caption{Overview of FlashSVD v1.5. We reorganize the fragmented, shattered low-rank execution path (left) into a streamlined, thin serving path (right) via graph replay, eliminating repeated launch bubbles and host bottlenecks.}
    \label{fig:placeholder}
\end{figure*}

\section{Related Work}

\paragraph{SVD for Checkpoint Compression.} 
While early works successfully compressed small encoders using standard SVD~\cite{chen2021drone,gao2024adarankenc,hsu2022fwsvd}, modern LLMs require more advanced treatments. Current open-source solutions typically fall into three directions. First, whitening-based methods like ASVD and SVD-LLM apply activation-aware scaling prior to decomposition to ensure truncation is aligned with model performance on down-streaming tasks~\cite{yuan2023asvd,wang2025svdllm,gao2026dfsvd,gao2025gfsvd,hu2026saes,ding2025dipsvd,lin2024modegpt,wang2025svdllmv2,wang2025dobisvd,li2025adasvd,mi2025drank,chekalina2025gfwsvd,sy2025liLLaMA}. Second, the activation-space truncation methods such as Dobi-SVD, aggressively fit models into extrememly low-ranked activation spaces by exploring the optimal activation subspace in backward propagation and project weights into that low-rank subspace, along with combining SVD with mixed-precision quantization~\cite{wang2025dobi,sinha2026aasvd,abbasi2026zssvd}. However, since our goal is to evaluate pure low-rank execution, we consider these quantization-hybrid methods orthogonal to our scope and only consider the activation truncation design and assume strictly uniform-precision baselines. Third, parameter-sharing techniques~\cite{wang2024basissharing, mi2025drank} have recently emerged as a highly effective approach. Instead of maintaining independent decomposition factors for each layer, these methods explore a global, shared set of low-rank bases that is reused across multiple Transformer blocks. By decoupling this shared structural basis from layer-specific projections, they drastically reduce the parameter footprint while maintaining model capacity.


\paragraph{Attention-Only Decomposition.} 
Parallel to checkpoint compression, another line of research focuses entirely on runtime activations. Methods such as Palu, xKV, and QSVD~\cite{chang2025palu,chang2025xkv,wang2025qsvd,saxena2024eigen} apply SVD exclusively to the KV cache while leaving the MLP layers completely uncompressed. This strategy is highly effective for resource-rich, long-context serving where the dynamic KV cache is the absolute memory bottleneck. Our target scenario, however, is fundamentally different. In edge deployments—which are characterized by short contexts and small batch sizes—the static model weights consume the vast majority of memory. Therefore, FlashSVD v1.5 deliberately targets checkpoint compression rather than activation compression, directly tackling the true bottleneck of edge inference.

\begin{figure*}
    \centering
    \begin{minipage}{0.49\linewidth}
        \centering
        \includegraphics[width=\linewidth]{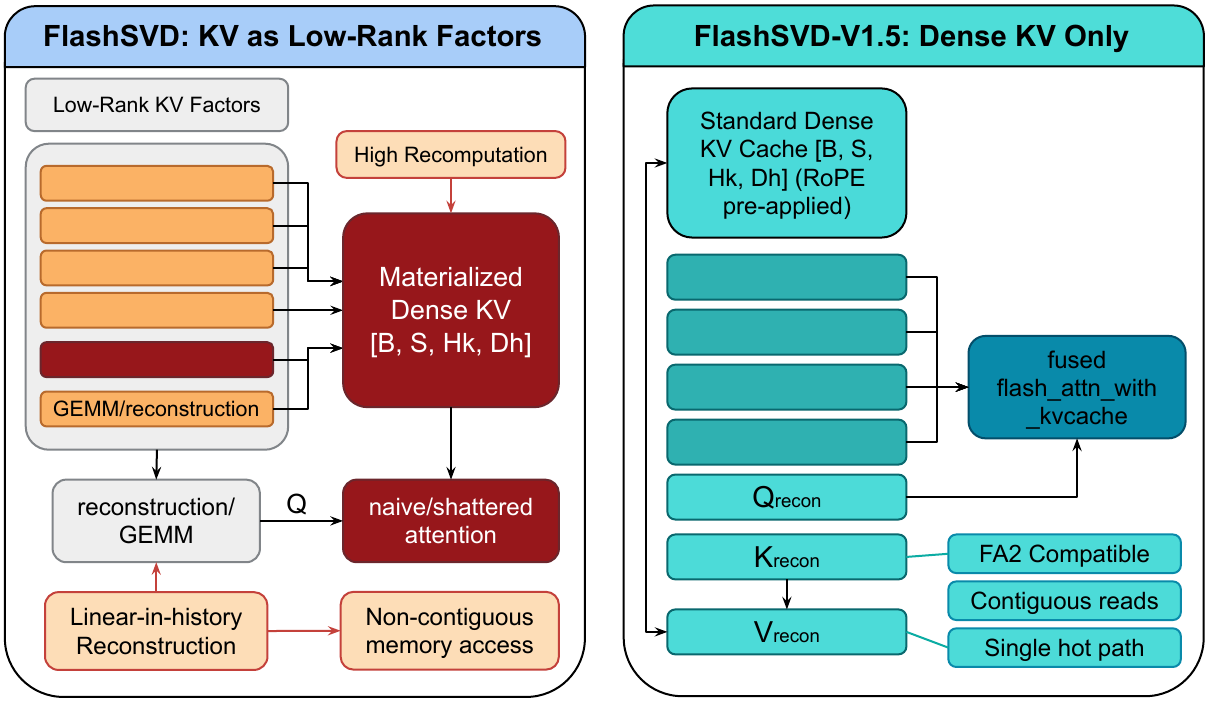}
        \caption{Transition from low-rank KV factors to dense-KV attention. By materializing historical KV states into a contiguous dense cache, FlashSVD v1.5 avoids the high recomputation and irregular memory access overhead of previous low-rank inference methods.}
        \label{fig:attn_placeholder}
    \end{minipage}\hfill
    \begin{minipage}{0.49\linewidth}
        \centering
        \includegraphics[width=\linewidth]{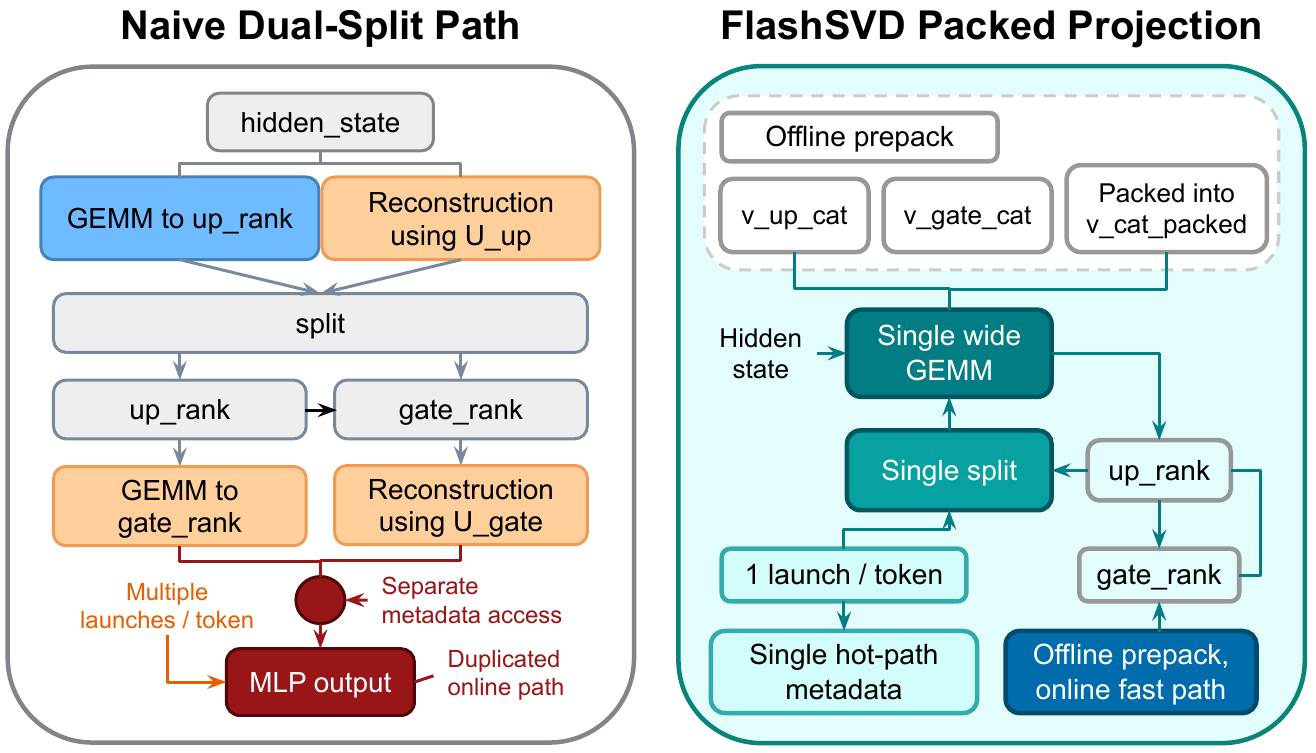}
        \caption{Reducing kernel fragmentation in MLP blocks. Instead of a dual-split execution, FlashSVD v1.5 utilizes offline prepacking and online packed projection to merge two duplicated paths into one, significantly reducing launch overhead.}
        \label{fig:mlp_placeholder}
    \end{minipage}
\end{figure*}

\section{Methodology}

FlashSVD v1.5 targets latency-sensitive small-batch LLM serving, especially autoregressive decoding at batch size $B=1$. In this regime, low-rank checkpoints often fail to realize their theoretical efficiency not because they lack arithmetic savings, but because the runtime path becomes fragmented. As illustrated in Figure~1, a naive low-rank implementation breaks a Transformer layer into many short-lived projections, reconstructions, cache-handling steps, and kernel launches, creating repeated host interaction and launch bubbles. FlashSVD v1.5 is designed around a simple principle: rather than exposing this shattered execution path directly, we reorganize it into a thin serving path. Concretely, the runtime combines phase-specific execution with three decode-oriented components: dense-KV attention, packed MLP projection, and per-layer graph replay.

\subsection{FlashSVD v1.5 Design}

\paragraph{Unified runtime across checkpoint families.}
Before execution, FlashSVD v1.5 maps different low-rank checkpoint families, including SVD-LLM v1, SVD-LLM v2, Dobi-SVD, and Basis Sharing, into a common native factorized representation. This normalization allows all supported checkpoints to enter the same runtime and use the same prefill kernels, decode fast paths, and packed-factor caches. For Basis Sharing, we additionally restore true shared \texttt{Parameter} objects across grouped layers so that shared bases remain physically shared and can be packed and reused once.

\paragraph{Attention: reconstruct the present once, read the past contiguously.}
The main bottleneck of low-rank decode attention is that the runtime keeps paying to reconstruct or reinterpret the history. Figure~2 contrasts the two execution styles. In the naive path, historical KV states remain exposed as low-rank factors, which introduces repeated reconstruction, irregular memory access, and a fragmented attention hot path. This shattered execution with heavy reconstruction explains the bottleneck in the original FlashSVD, where the latency penalty of repeated reconstruction outweighs the marginal memory benefits. FlashSVD v1.5 instead reconstructs only the current-token dense \texttt{q/k/v}, stores historical \texttt{K/V} in a standard dense KV cache compatible with \texttt{flash\_attn\_with\_kvcache}, and executes the history-dependent attention step through one FA2-friendly path. Intuitively, we reconstruct the present once and read the past in a contiguous layout, instead of repeatedly reconstructing the past at every decode step. During prefill, where full-sequence execution remains favorable, FlashSVD v1.5 still uses a dedicated factorized attention kernel and avoids explicit dense \texttt{Q/K/V} materialization.

\paragraph{MLP: merge two duplicated online paths into one packed path.}
Figure~3 shows a similar source of fragmentation in the MLP block. In a naive dual-split low-rank execution, the same hidden state is sent through separate input-side projections for the \texttt{up} and \texttt{gate} branches, which duplicates launches, metadata traffic, and online bookkeeping. FlashSVD v1.5 removes this duplication by offline-packing the compatible input-side factors of the \texttt{up} and \texttt{gate} branches into a single tensor, performing one wide projection online, and splitting the resulting rank activations afterward. The output-side reconstructions and down projection remain unchanged. As a result, the low-rank checkpoint is still executed exactly, but the online MLP path becomes much thinner.

\paragraph{Per-layer graph replay.}
These operator-level changes shorten the attention and MLP hot paths, but they do not by themselves remove all decode-time fragmentation. Even after dense-KV attention and packed MLP projection are introduced, an eager runtime still pays for many small boundaries between kernels and host launches. FlashSVD v1.5 therefore captures the stable small-batch decode body of each layer as a replayable CUDA graph. This turns the per-layer decode path into a compact replay unit and is the final step that converts low-rank arithmetic savings into end-to-end latency gains. In the terminology of Figure~1, the resulting runtime is composed of dense-KV attention, a merged low-rank path, and per-layer graph replay, rather than a collection of shattered low-rank fragments.

\section{Experiment}

\sisetup{
  detect-weight=true,
  detect-inline-weight=math,
  table-number-alignment=center
}

\begin{table*}[t]
\centering
\small
\setlength{\tabcolsep}{4.5pt}
\renewcommand{\arraystretch}{1.15}
\begin{threeparttable}
\caption{Main serving results of FlashSVD across representative prompt lengths. Decode latency is reported in ms/token, and end-to-end latency accounts for both the prefill and decode phases. Speedups are calculated using median values under identical precision, hardware, and serving configurations. } 
\label{tab:main}

\begin{tabular}{
@{} l
S[table-format=4.0]
S[table-format=3.0]
S[table-format=2.2]
S[table-format=2.2]
c
S[table-format=1.2]
S[table-format=1.2]
c @{}
}
\toprule
& \multicolumn{2}{c}{Sequence length}
& \multicolumn{3}{c}{Decode latency (ms/token)}
& \multicolumn{3}{c}{End-to-end latency (s)} \\
\cmidrule(lr){2-3}
\cmidrule(lr){4-6}
\cmidrule(l){7-9}
Baseline
& {Prompt}
& {Gen}
& {Base}
& {FlashSVD v1.5}
& {Speedup}
& {Base}
& {FlashSVD v1.5}
& {Speedup} \\
\midrule

\multirow{4}{*}{\begin{tabular}[c]{@{}l@{}}HF StaticCache \\ (SDPA) baseline\end{tabular}}
& 512  & 32  & 30.84 & 12.16 & \textbf{2.55$\times$} & 1.03 & 0.43 & \textbf{2.39$\times$} \\
& 2048 & 128 & 30.63 & 12.24 & \textbf{2.50$\times$} & 4.10 & 1.72 & \textbf{2.38$\times$} \\
& 4096 & 128 & 30.60 & 12.85 & \textbf{2.38$\times$} & 4.29 & 1.93 & \textbf{2.23$\times$} \\
& 8192 & 128 & 30.65 & 14.09 & \textbf{2.18$\times$} & 4.85 & 2.39 & \textbf{2.03$\times$} \\

\addlinespace[0.35em]

\multirow{4}{*}{\begin{tabular}[c]{@{}l@{}}Dense KV-Cache \\ + FA2 baseline\end{tabular}}
& 512  & 32  & 26.90 & 12.16 & \textbf{2.20$\times$} & 0.91 & 0.43 & \textbf{2.07$\times$} \\
& 2048 & 128 & 26.19 & 12.24 & \textbf{2.13$\times$} & 3.51 & 1.72 & \textbf{2.03$\times$} \\
& 4096 & 128 & 26.52 & 12.85 & \textbf{2.07$\times$} & 3.68 & 1.93 & \textbf{1.91$\times$} \\
& 8192 & 128 & 26.21 & 14.09 & \textbf{1.86$\times$} & 3.94 & 2.39 & \textbf{1.65$\times$} \\

\bottomrule
\end{tabular}

\end{threeparttable}
\end{table*}




\begin{figure*}[t]
    \centering
    \includegraphics[width=0.95\linewidth]{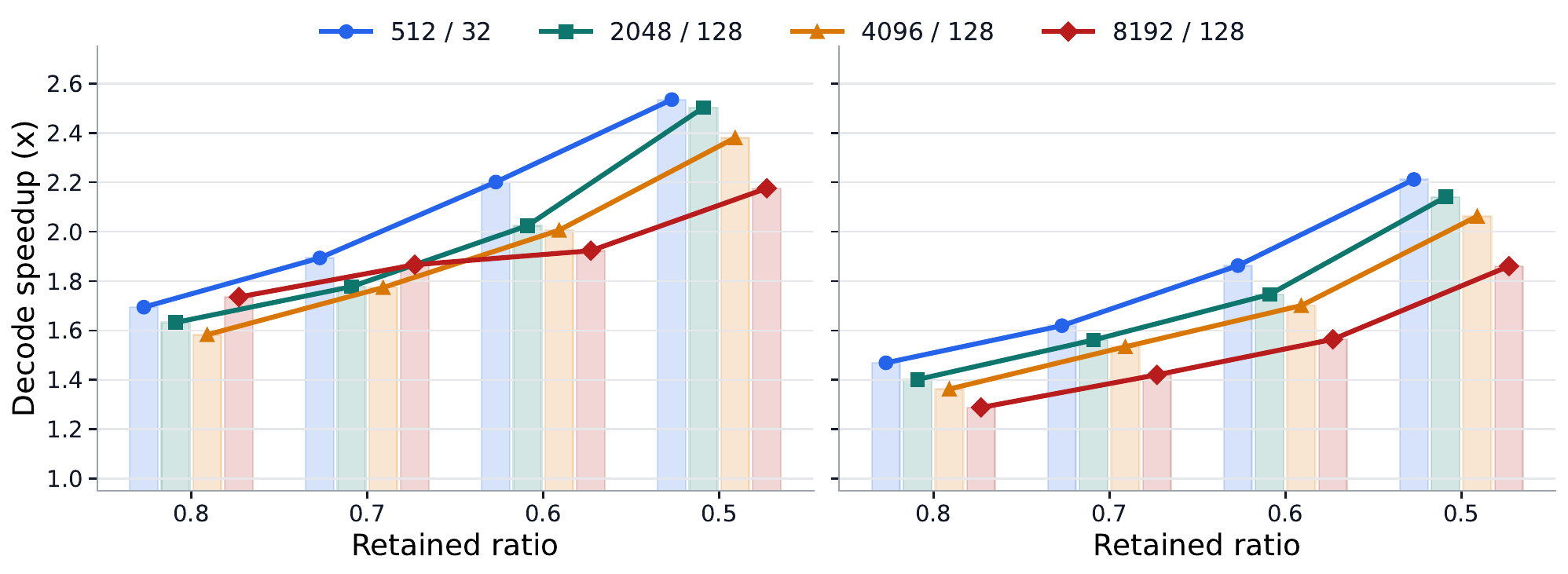}
    \caption{Robustness to retained ratio. Left: decode speedup of FlashSVD v1.5 over HF StaticCache across retained ratios and prompt settings. Right: the same comparison against the stronger Dense KV-Cache + FA2 baseline. As more rank is retained, the speedup decreases monotonically, but FlashSVD v1.5 remains consistently above $1\times$ even in the mild-compression regime.}
    \label{fig:ratio_robustness}
\end{figure*}

\subsection{Experimental Setup}

\paragraph{Baselines and evaluation goal.}
Our goal is to isolate the runtime contribution of FlashSVD v1.5 while keeping the compressed checkpoints fixed. We therefore compare FlashSVD v1.5 against two matched end-to-end baselines. The first is \textbf{HF StaticCache}, a practical low-rank serving baseline built on the standard HuggingFace-style static KV-cache runtime. The second is \textbf{Dense KV-Cache + FA2}, a stronger aligned baseline that reconstructs dense Q/K/V from the same compressed checkpoints, applies external RoPE, and performs decode with \texttt{flash\_attn\_with\_kvcache}. Unless otherwise stated, all methods are evaluated under identical checkpoints, precision, and hardware settings. Historical comparisons to FlashSVD v1 and additional encoder results are deferred to the appendix.

\paragraph{Checkpoint families, workloads, and metrics.}
We evaluate FlashSVD v1.5 on representative public low-rank checkpoint families, including \textbf{SVD-LLM v1}, \textbf{SVD-LLM v2}, and \textbf{Basis Sharing}, under a common aligned LLaMA-7B serving recipe. Our primary target is latency-sensitive autoregressive decoder serving at batch size $B=1$. We report decode latency, end-to-end generation latency, and repeated-run statistics where available. Beyond the main short- and medium-context settings, we additionally evaluate retained-ratio robustness, long-generation behavior, and decoder-side mechanism analyses to test whether FlashSVD v1.5 improves the actual token-serving path rather than merely reducing theoretical FLOPs. \textbf{Evaluation philosophy.} Our evaluation is deliberately decoder-centric. The main question is not whether low-rank compression reduces arithmetic in principle, but whether a serving runtime can convert fixed low-rank checkpoints into faster token generation in practice. We therefore emphasize matched serving settings, decoder-side latency measurements, and mechanism-oriented ablations, while deferring lower-level implementation checks and fidelity audits to supporting discussion.


\subsection{Main Results}

\paragraph{Main serving results.}
Table~\ref{tab:main} is the primary decoder result. Across representative serving settings, FlashSVD v1.5 consistently improves both decode latency and end-to-end latency relative to HF StaticCache. The same ordering also holds against the stronger Dense KV-Cache + FA2 baseline, showing that the gain is not an artifact of a weak reference path. As context length increases, the relative speedup narrows, but remains clearly positive throughout the evaluated range. Overall, the table shows that FlashSVD v1.5 converts low-rank checkpoints into real wall-clock serving gains under matched conditions.

\paragraph{Cross-family generality.}
Table~\ref{tab:family_coverage} shows that these gains are not tied to a single checkpoint export format. Across SVD-LLM v1, SVD-LLM v2, and Basis Sharing, the decode-side improvement is consistent, supporting the claim that the unified runtime abstraction transfers across native factorized checkpoints, wrapper-exported checkpoints, and basis-sharing variants. Prefill behavior is more family-dependent, but the decode-side trend remains stable across the full sweep.

\begin{figure*}[t]
    \centering
    \includegraphics[width=0.95\linewidth]{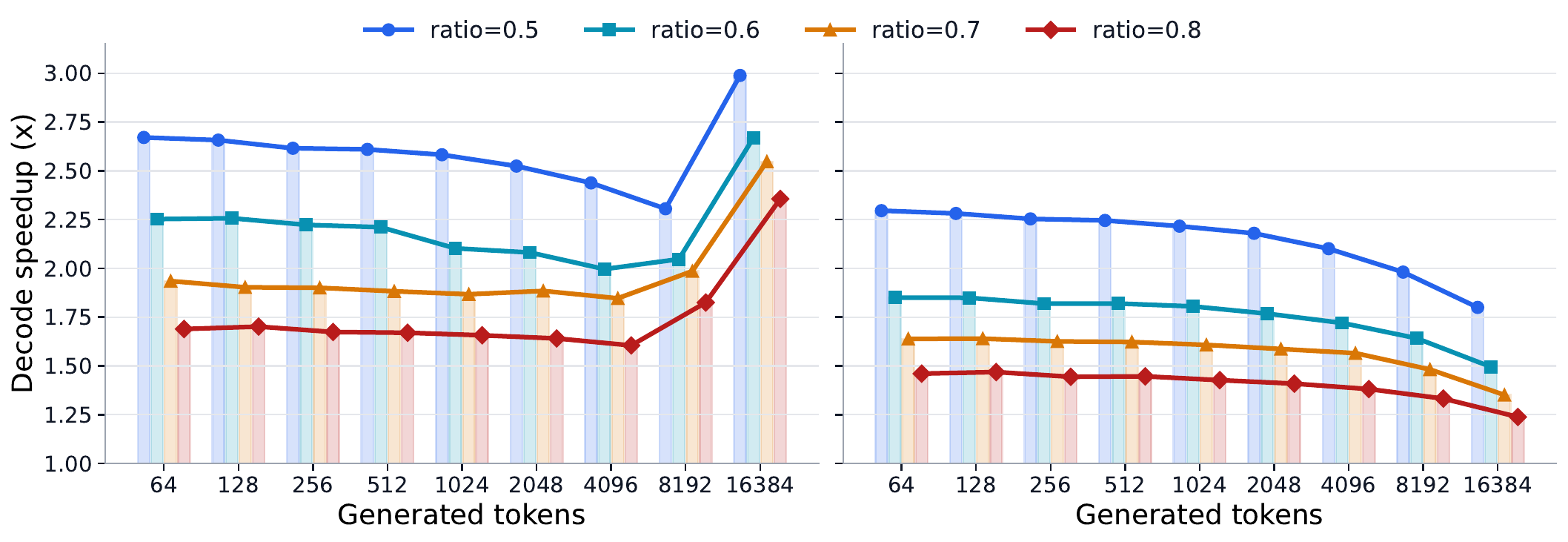}
    \caption{Robustness over long generations. Left: decode speedup of FlashSVD v1.5 over HF StaticCache as the generated length increases from 64 to 16{,}384 tokens under retained ratios 0.5--0.8 with prompt length 512. Right: the same comparison against Dense KV-Cache + FA2. As context length increases, standard SDPA becomes bottlenecked by memory bandwidth. FlashSVD v1.5 avoids this by enabling FA2-compatible dense-KV reads, which leads to higher speedups during long steady-state generation.}
    \label{fig:long_generation_robustness}
\end{figure*}

\begin{table}[t]
\centering
\small
\setlength{\tabcolsep}{4.8pt}
\renewcommand{\arraystretch}{1.05}
\begin{threeparttable}
\caption{Cross-family coverage of FlashSVD v1.5 on public SVD families under the unified LLaMA-7B serving recipe.}
\label{tab:family_coverage}
\begin{tabular}{@{}lrrrr@{}}
\toprule
Family & \# ckpts & Prefill spd. & Decode spd. & E2E spd. \\
\midrule
SVD-LLM v1    & 5  & 1.55$\times$ & 1.45$\times$ & 1.46$\times$ \\
SVD-LLM v2    & 5  & 1.14$\times$ & 1.50$\times$ & 1.45$\times$ \\
Basis Sharing & 3  & 0.91$\times$ & 1.49$\times$ & 1.42$\times$ \\
\midrule
Overall       & 13 & 1.25$\times$ & 1.48$\times$ & 1.44$\times$ \\
\bottomrule
\end{tabular}
\end{threeparttable}
\end{table}

\paragraph{Robustness to retained ratio.}
Figure~\ref{fig:ratio_robustness} shows that the decoder-side gain persists even as compression becomes milder. As more rank is retained, the relative speedup decreases, which is expected because the factorized operators move closer to the dense model and expose less redundancy for the runtime to exploit. Even so, FlashSVD v1.5 remains beneficial throughout the evaluated range. This result is important for deployment: the runtime helps not only in aggressively compressed settings, but also in conservative regimes that stay closer to lossless behavior.

\paragraph{Robustness over long generations.}
Figure~\ref{fig:long_generation_robustness} shows that the benefit persists throughout long decode horizons. The advantage remains visible across retained ratios and across the full generated-length sweep, indicating that FlashSVD v1.5 improves the steady-state token-serving path rather than only reducing one-time startup overhead.


\section{Decoder Mechanism Analysis}

\begin{figure*}[t]
    \centering
    \begin{minipage}{0.48\linewidth}
        \centering
        \includegraphics[height=4.6cm,keepaspectratio]{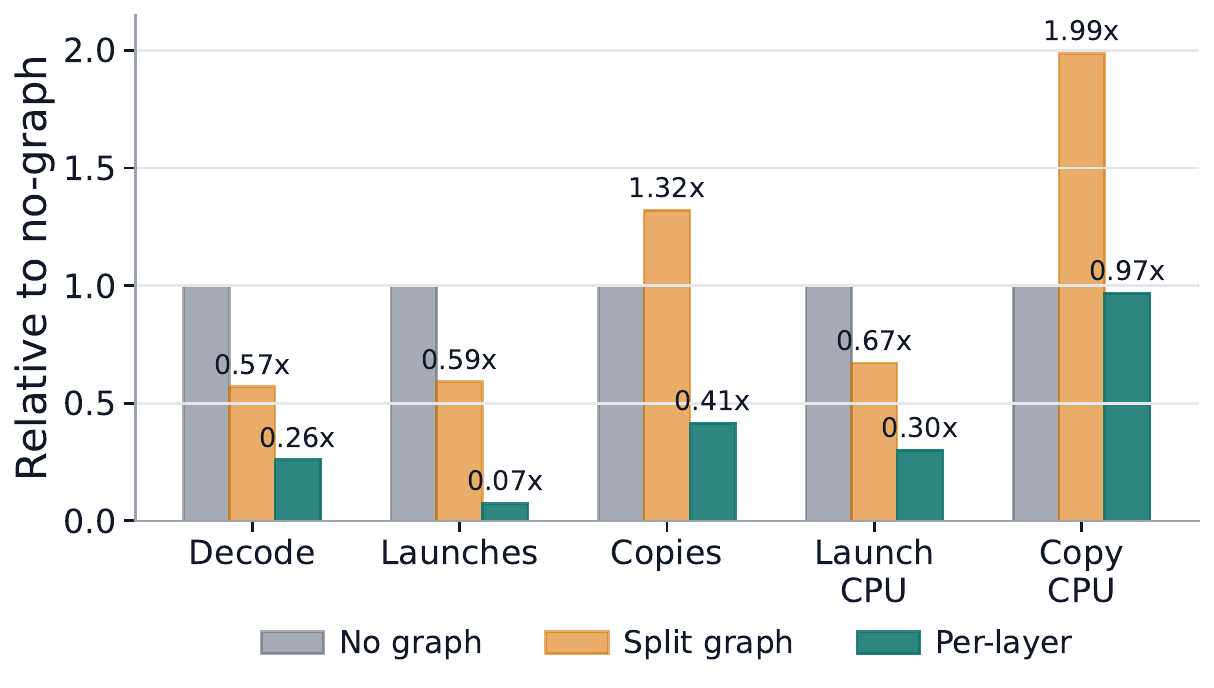}
        \caption{Graph-replay micro-ablation, normalized to the no-graph execution. Partial graphing reduces some launches, but can also increase copy traffic and copy-side CPU overhead. Per-layer replay simultaneously reduces decode time, launch count, and CPU launch overhead, showing that the graph boundary must be coarse enough to eliminate fragmentation rather than relocate it.}
        \label{fig:graph_ablation}
    \end{minipage}\hfill
    \begin{minipage}{0.48\linewidth}
        \centering
        \includegraphics[height=4.6cm,keepaspectratio]{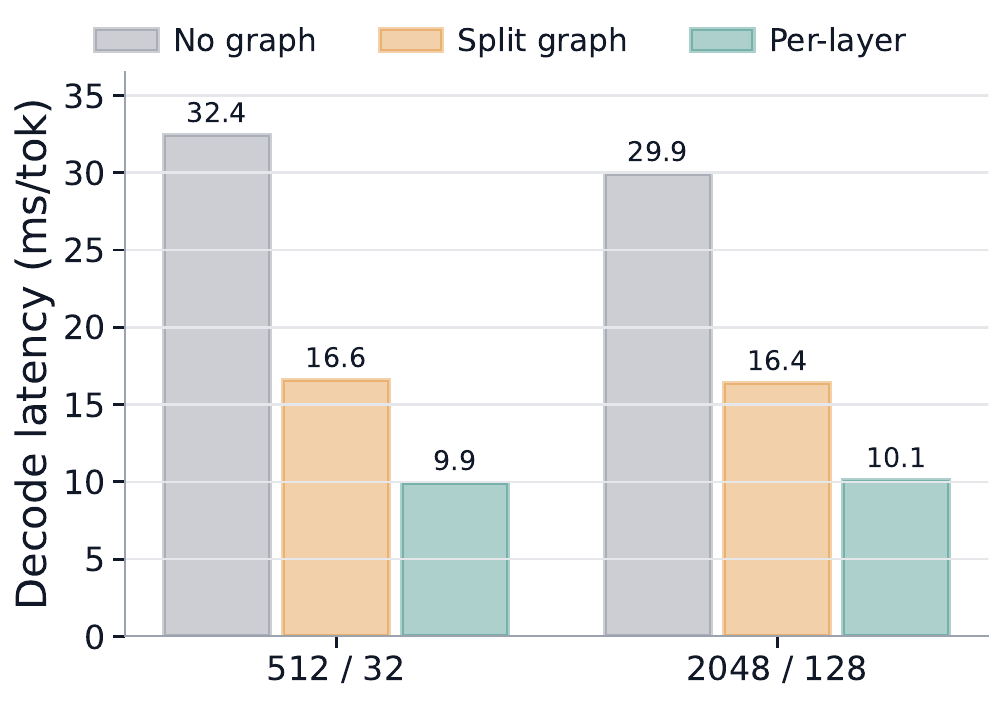}
        \caption{Ablation of graph granularity on absolute decode latency. Capturing the decode body as a per-layer replay unit outperforms both eager execution and split-graph execution at short and medium contexts, confirming that coarse-grained replay is the correct granularity for low-rank decode serving.}
        \label{fig:graph_granularity}
    \end{minipage}
\end{figure*}

\paragraph{Per-layer graph replay is the right decode unit.}
Figures~\ref{fig:graph_ablation} and \ref{fig:graph_granularity} together isolate the role of graph replay in the decoder path. The micro-ablation shows that partial graphing removes some launch overhead but also relocates work into graph-boundary traffic, whereas per-layer replay improves the relevant runtime counters jointly. The latency ablation then confirms the same conclusion in end-to-end decode time: per-layer replay consistently outperforms both eager execution and split-graph execution. Together, these results show that the main benefit does not come from graph capture in the abstract, but from capturing the right decode body.

\begin{figure}[ht]
    \centering
    \includegraphics[width=0.75\linewidth]{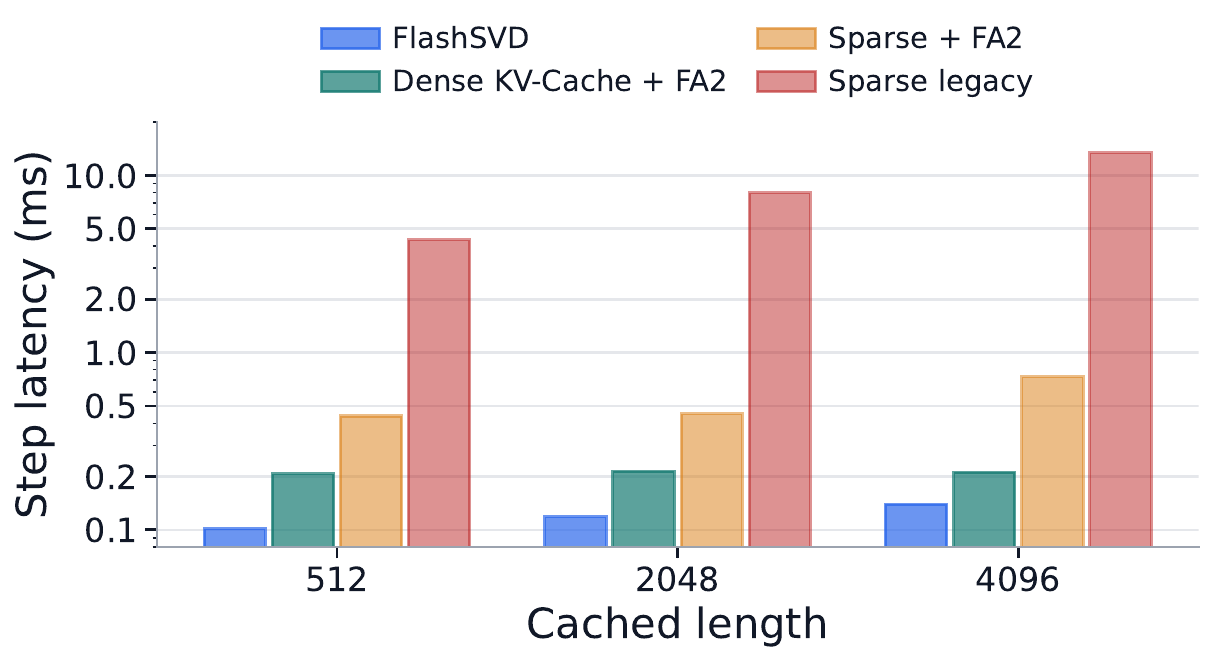}
    \caption{Attention-route ablation across cached lengths. Keeping historical KV in a dense FA2-compatible layout yields substantially flatter step-latency scaling than sparse history paths. FlashSVD achieves the lowest latency by combining this dense-KV attention route with the rest of the thin serving path.}
    \label{fig:attention_route}
\end{figure}

\paragraph{Dense-KV is the right decode-time attention route.}
Figure~\ref{fig:attention_route} isolates the attention path as cached length increases. Keeping historical KV in a dense FA2-compatible layout yields much flatter scaling than routes that retain history in low-rank form. At the same time, FlashSVD remains faster than Dense KV-Cache + FA2 across the sweep, showing that dense-KV attention is necessary but not sufficient: the remaining gain comes from integrating that route into a thinner end-to-end serving path.


\paragraph{Fidelity and Microbenchmark Validation.} 
To ensure our system-level optimizations do not degrade generation quality, we conduct a strict decoder-side fidelity audit against standard fp32 and bf16 baselines. FlashSVD v1.5 tightly tracks the practical behavior of HuggingFace StaticCache under bf16 execution, maintaining perfect first-token agreement with fp32 references and substantial pairwise agreement throughout long generation trajectories. Additionally, local microbenchmarks confirm that our packed FFN serving route activates precisely as intended without introducing unintended computation overhead. Detailed fidelity audits and local routing latencies are deferred to Appendix~\ref{app:decoder_appendix}.

\paragraph{Encoder-side results.}
Encoder experiments show that the systems benefits of FlashSVD extend beyond autoregressive decoding. Across representative compressed BERT workloads, \texttt{flashsvd15} is the strongest throughput- and latency-oriented backend, while both FlashSVD backends substantially reduce peak memory relative to the naive and \texttt{sdpa} baselines. This ordering is consistent across representative task-level comparisons and persists as sequence length and batch size increase, indicating that the gain is a backend-level effect rather than a task-specific artifact. The original \texttt{flashsvd} remains slightly more memory-minimal, whereas \texttt{flashsvd15} provides the best overall operating point when both throughput and memory efficiency matter.

\paragraph{Why the encoder gains arise.}
The appendix further shows that these encoder-side improvements arise from the same systems mechanisms emphasized in the decoder setting. Compression alone reduces parameter storage, but naive low-rank execution still incurs substantial activation and rank-expansion overhead. The FlashSVD backends reduce these costs by eliminating intermediate attention-related buffers and by consolidating fragmented GEMM dispatch into fused execution paths. The encoder Pareto views also clarify the separation between checkpoint quality and runtime efficiency: the compressed checkpoint determines the retained task accuracy, while the backend determines how efficiently that checkpoint is served. Taken together, the encoder results reinforce the same broader conclusion as the decoder results: practical low-rank acceleration depends on runtime co-design, not compression alone. Detailed encoder backend comparisons, scaling studies, profiling analyses, and Pareto views are deferred to Appendix Table~\ref{tab:svd_backend_tasks_synthetic} and Figures~\ref{fig:fig07_seqlen_combined}, \ref{fig:fig08_batch_combined}, \ref{fig:memory_breakdown_appendix}, \ref{fig:fig15_nsys_combined}, and \ref{fig:pareto_appendix}.

\section{Conclusion}

We presented FlashSVD v1.5, a unified inference runtime for low-rank transformer checkpoints. The main result of the paper is that low-rank checkpoints do not automatically translate into fast serving: practical speedups require a thinner runtime path. Across fixed low-rank decoder checkpoints, FlashSVD v1.5 consistently improves decode and end-to-end latency, and these gains transfer across checkpoint families, compression regimes, and long decode horizons. Mechanistically, the results show that dense-KV attention and per-layer graph replay are the key ingredients for converting low-rank arithmetic savings into wall-clock decoder speedups.

More broadly, our results suggest that low-rank compression should be evaluated jointly with systems support rather than in isolation. FlashSVD v1.5 focuses primarily on latency-sensitive decoder serving under practical bf16 cached execution, where it remains close to the baseline cached decode path without claiming token-by-token identity. Extending the same runtime principles to broader serving regimes and strengthening encoder-side support are natural directions for future work.

\bibliography{text/refs}
\bibliographystyle{unsrt} 

\newpage
\appendix
\onecolumn

\paragraph{Why we do not use FlashSVD as a primary decoder baseline.}
Although FlashSVD v1 is the historical predecessor of FlashSVD v1.5, we do not use it as a primary headline baseline in the main tables. The reason is that FlashSVD v1 and FlashSVD v1.5 do not represent the same end-to-end serving stack under a common evaluation recipe: the earlier codebase includes heterogeneous legacy encoder-oriented and archived decoder paths, whereas FlashSVD v1.5 is evaluated as a unified low-rank decoder-serving runtime on fixed compressed checkpoints. To isolate the runtime contribution of FlashSVD v1.5 under matched checkpoint, precision, and hardware settings, we therefore use \textbf{StaticCache} and \textbf{DenseKVCacheBaseline} as the main baselines. Historical comparisons to earlier FlashSVD-style decode components are better interpreted as ablation or lineage evidence rather than as the primary end-to-end baseline.

\section{Additional Decoder Evidence}
\label{app:decoder_appendix}

This appendix provides two decoder-side supporting checks that complement the main paper. First, we validate the local FFN path used by the packed MLP serving routine. Second, we report a decoder-side fidelity audit against a common fp32 no-cache reference. These results should be read as supporting evidence for the decoder story rather than as additional headline results.

\subsection{Local FFN-Path Validation}




Table~\ref{tab:decoder_ffn_local_latency} verifies the default FFN backend-selection policy. Without graph replay, the runtime stays on the eager no-merge path; with layer-tail graph replay enabled, it switches to the merged backend. In both cases, the selected backend matches the corresponding explicit configuration exactly, and explicit user overrides are preserved. This table should therefore be read as a backend-selection correctness check rather than a speed comparison.

\begin{table}[t]
\centering
\small
\begin{tabular}{@{}lr@{}}
\toprule
Configuration & Mean latency (ms) \\
\midrule
\texttt{no\_merge eager} & 0.2662 \\
\texttt{auto} + no graph & 0.2608 \\
Explicit \texttt{prod + layer\_tail\_graph} & 0.1921 \\
\texttt{auto} + \texttt{layer\_tail\_graph} & 0.1921 \\
\bottomrule
\end{tabular}
\caption{Local FFN-path latency under routed and explicit configurations. The automatic route stays close to the eager no-graph path when graph replay is disabled, and exactly matches the explicit merged configuration when layer-tail graph replay is enabled.}
\label{tab:decoder_ffn_local_latency}
\end{table}

Table~\ref{tab:decoder_ffn_local_latency} provides local latency support for the packed FFN path. In the graph-disabled setting, the automatic route remains close to the eager no-merge path, showing that the default policy does not activate the merged backend prematurely. With layer-tail graph replay enabled, the automatic route matches the explicit merged configuration, indicating that the intended FFN fast path is used only in the graph-enabled regime. We therefore interpret this table as supporting evidence for the packed FFN mechanism, not as a claim that the FFN path alone explains the full decoder speedup.

\subsection{Decoder-Side Fidelity Audit}

\begin{table}[t]
\centering
\small
\begin{tabular}{@{}lccc@{}}
\toprule
System & \shortstack[c]{Exact match\\vs.\ FP32 gold} & \shortstack[c]{First-token\\match} & \shortstack[c]{Mean token\\match} \\
\midrule
\multicolumn{4}{@{}l}{\textit{vs.\ FP32 no-cache reference}} \\
\midrule
\shortstack[l]{HF StaticCache\\(bf16)} & 14 / 20 & 20 / 20 & 0.8070 \\
\shortstack[l]{FlashSVD-v1.5 prod\\(bf16)} & 13 / 20 & 20 / 20 & 0.7461 \\
\midrule
\multicolumn{4}{@{}l}{\textit{Pairwise bf16 cached agreement}} \\
\midrule
\shortstack[l]{HF StaticCache\\vs.\ FlashSVD-v1.5} & \multicolumn{3}{l}{Exact agreement on 13 / 20 prompts} \\
\bottomrule
\end{tabular}
\caption{Greedy decode fidelity on 20 prompts with 64 generated tokens. The reference is an fp32 no-cache decode. Both cached bf16 systems match the fp32 reference on the first generated token for all audited prompts, but neither is token-by-token identical to the fp32 gold trajectory.}
\label{tab:decoder_fidelity_audit}
\end{table}

Table~\ref{tab:decoder_fidelity_audit} should be read as a decode-fidelity audit rather than a claim of exact equivalence. Against a common fp32 no-cache reference, both practical bf16 cached paths recover the first generated token on all audited prompts, indicating stable early-step behavior. Over the full decode trajectory, the HuggingFace StaticCache path is slightly closer to the fp32 gold in this audit, while FlashSVD still remains close to the baseline cached decode path and shows substantial pairwise agreement with it. We therefore use this table to support the claim that FlashSVD tracks practical cached decoding behavior without claiming token-by-token identity under bf16 cached execution.

\section{Additional Encoder Experiments}

This appendix provides the encoder-side evidence underlying the summary in the main text. We focus on four questions. First, what is the backend ordering under representative compressed BERT workloads? Second, how do these differences scale with sequence length and batch size? Third, how should these runtime effects be interpreted relative to checkpoint quality? Finally, what systems mechanisms produce the observed gains? The following results show that the same runtime principles emphasized in the decoder setting also apply to encoder inference: fused execution improves throughput, FlashSVD backends substantially reduce memory, and the resulting efficiency gains become more pronounced as the workload shifts toward the activation-dominated regime.

\subsection{Backend Summary Across Representative Tasks}

\begin{table}[!t]
\centering
\small
\setlength{\tabcolsep}{4pt}
\renewcommand{\arraystretch}{1.10}
\begin{threeparttable}
\caption{
Backend comparison for SVD-compressed BERT on representative GLUE tasks under synthetic inputs.
Results are reported on MNLI, QQP, and STS-B using bf16, sequence length 512, and batch size 32.
Synthetic inputs use fixed full-length token sequences, so they remove dataset-specific sequence-length variation and padding effects.
We compare four inference backends (\texttt{naive}, \texttt{sdpa}, \texttt{flashsvd}, and \texttt{flashsvd15}) in terms of latency, throughput, and peak GPU memory.
Lower latency and memory are better, while higher throughput is better.
}
\label{tab:svd_backend_tasks_synthetic}
\begin{tabular}{llccc}
\toprule
Task & Backend & Latency (ms) $\downarrow$ & Throughput $\uparrow$ & Peak Mem. (MB) $\downarrow$ \\
\midrule
\multirow{4}{*}{MNLI}
& naive      & 51.34 & 623.3  & 989.2 \\
& sdpa       & 25.33 & 1263.1 & 605.2 \\
& flashsvd   & 44.40 & 720.8  & \textbf{341.4} \\
& flashsvd15 & \textbf{22.52} & \textbf{1421.0} & 343.9 \\
\midrule
\multirow{4}{*}{QQP}
& naive      & 51.37 & 622.9  & 989.2 \\
& sdpa       & 25.48 & 1255.8 & 605.2 \\
& flashsvd   & 44.67 & 716.3  & \textbf{341.4} \\
& flashsvd15 & \textbf{22.69} & \textbf{1410.6} & 345.4 \\
\midrule
\multirow{4}{*}{STS-B}
& naive      & 51.41 & 622.4  & 989.2 \\
& sdpa       & 25.50 & 1254.8 & 605.2 \\
& flashsvd   & 44.86 & 713.4  & \textbf{341.4} \\
& flashsvd15 & \textbf{22.63} & \textbf{1414.2} & 343.9 \\
\bottomrule
\end{tabular}
\begin{tablenotes}[flushleft]
\footnotesize
\item Results are shown for SVD-compressed BERT under different backends with synthetic inputs.
\item The same overall trend as in real-input evaluation is preserved: \texttt{flashsvd15} achieves the best latency and throughput, while \texttt{flashsvd} yields slightly lower peak memory.
\end{tablenotes}
\end{threeparttable}
\end{table}

Table~\ref{tab:svd_backend_tasks_synthetic} provides the clearest summary of encoder-side backend behavior. Across representative task-level comparisons, the ordering is stable: \texttt{flashsvd15} is the strongest throughput- and latency-oriented backend, while both FlashSVD backends substantially reduce peak memory relative to the naive and \texttt{sdpa} baselines. The original \texttt{flashsvd} remains slightly more memory-minimal, whereas \texttt{flashsvd15} provides the best overall operating point when both throughput and memory efficiency matter. The near-identical ordering across MNLI, QQP, and STS-B indicates that these results are driven primarily by backend execution structure rather than by task-specific artifacts.

\FloatBarrier

\subsection{Scaling Behavior}

\begin{figure*}[!t]
    \centering
    \includegraphics[width=1\textwidth]{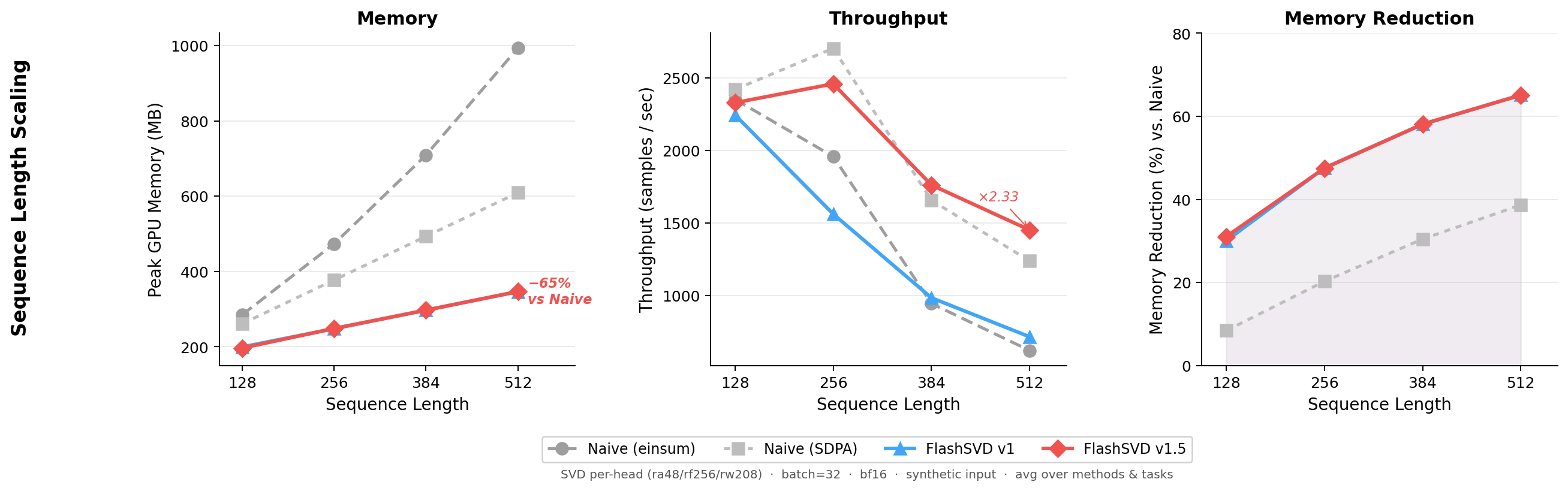}
    \caption{
    Sequence-length scaling for SVD-compressed BERT under different backends.
    From left to right, we report peak GPU memory, throughput, and relative memory reduction versus the naive backend.
    Results are collected at bf16 with batch size 32 using synthetic full-length inputs, and are averaged over compressed methods and tasks.
    FlashSVD backends scale more favorably with sequence length than the naive and \texttt{sdpa} baselines, with \texttt{flashsvd15} delivering the strongest throughput and the largest practical efficiency gains at longer sequence lengths.
    }
    \label{fig:fig07_seqlen_combined}
\end{figure*}

Figure~\ref{fig:fig07_seqlen_combined} shows that the encoder-side advantage of FlashSVD grows with sequence length. As inputs become longer, the memory savings widen and the throughput gap between \texttt{flashsvd15} and the baselines becomes more pronounced. This trend is consistent with the changing composition of the workload: at short sequence lengths, parameter storage remains a large fraction of total usage, whereas at longer sequence lengths, attention-related activation overhead becomes dominant. In that regime, fused execution is increasingly valuable because it avoids the intermediate buffers that generic low-rank backends still materialize.

\begin{figure}[!t]
    \centering
    \includegraphics[width=0.85\linewidth]{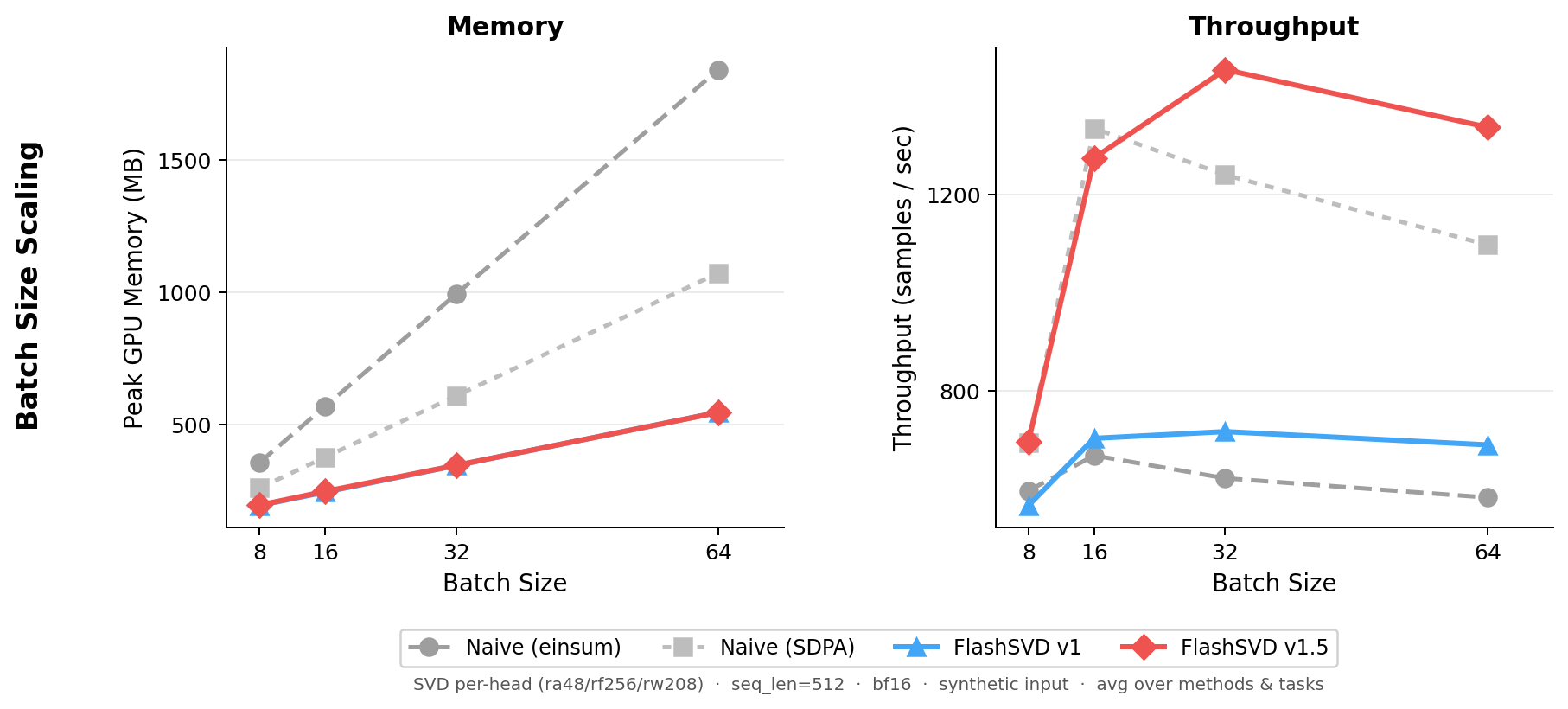}
    \caption{
    Batch-size scaling for SVD-compressed BERT under different backends.
    The left panel reports peak GPU memory, and the right panel reports throughput.
    Results are collected at bf16 with sequence length 512 using synthetic full-length inputs, and are averaged over compressed methods and tasks.
    FlashSVD backends substantially reduce memory across the full batch-size sweep, while \texttt{flashsvd15} delivers the strongest throughput at moderate and large batch sizes.
    }
    \label{fig:fig08_batch_combined}
\end{figure}

Figure~\ref{fig:fig08_batch_combined} shows the same conclusion from a different scaling axis. As batch size increases, all methods consume more memory, but the relative advantage of the FlashSVD backends remains large and the throughput benefit of \texttt{flashsvd15} becomes especially clear once the workload provides enough parallelism to keep the fused path busy. In contrast, the naive backend saturates earlier and becomes increasingly limited by activation overhead and dispatch inefficiency. Together, Figures~\ref{fig:fig07_seqlen_combined} and \ref{fig:fig08_batch_combined} show that the encoder-side gains are not confined to a single evaluation point; they strengthen as the workload becomes more demanding.

\FloatBarrier

\subsection{Accuracy--Efficiency View}

\begin{figure*}[!t] 
    \centering
    \begin{minipage}[t]{0.48\linewidth}
        \centering
        \includegraphics[width=\linewidth]{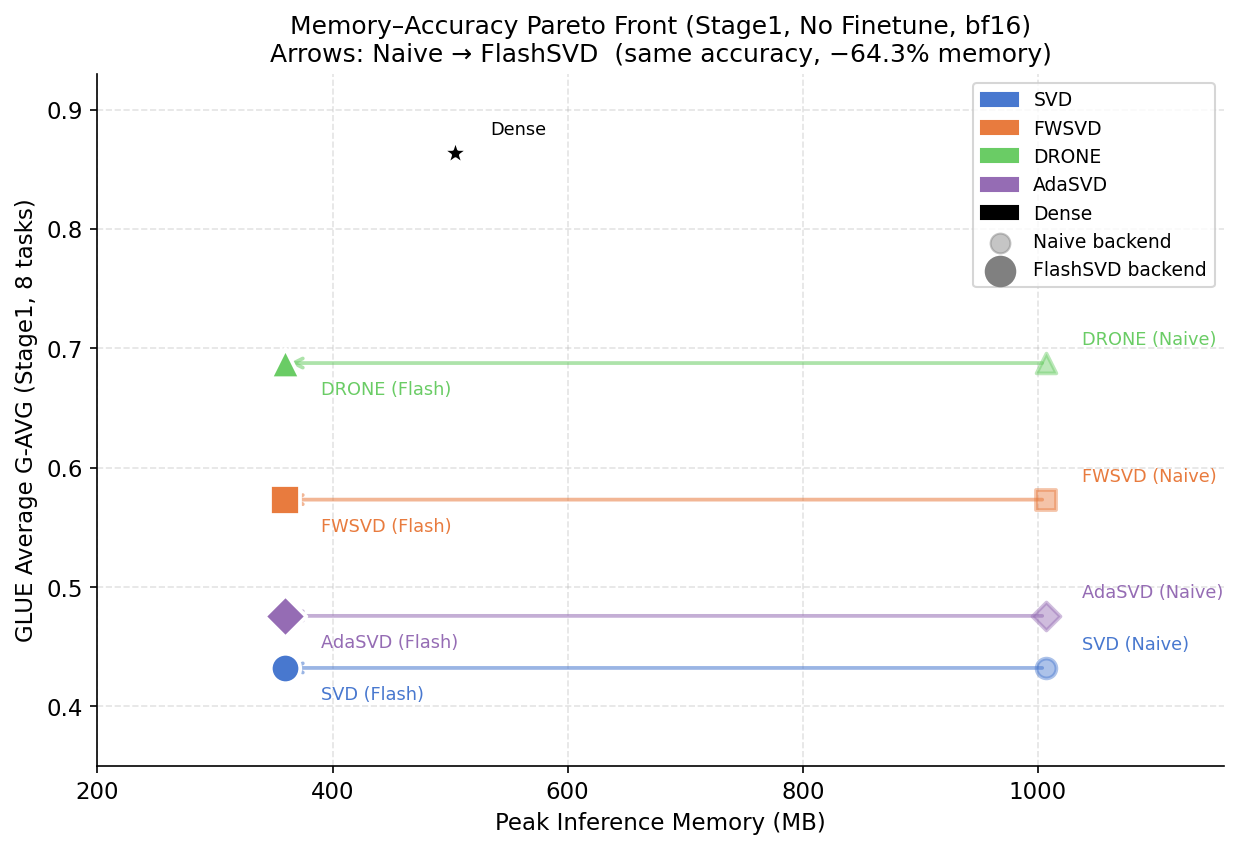}
        \caption{
        Memory--accuracy Pareto frontier for compressed BERT checkpoints under different backends.
        Results are shown for stage-1 evaluation at bf16.
        Each point compares peak inference memory against average GLUE accuracy, illustrating the trade-off between efficiency and retained task performance.
        Higher accuracy and lower memory are preferred.
        }
        \label{fig:pareto_appendix}
    \end{minipage}\hfill
    \begin{minipage}[t]{0.48\linewidth}
        \centering
        \includegraphics[width=\linewidth]{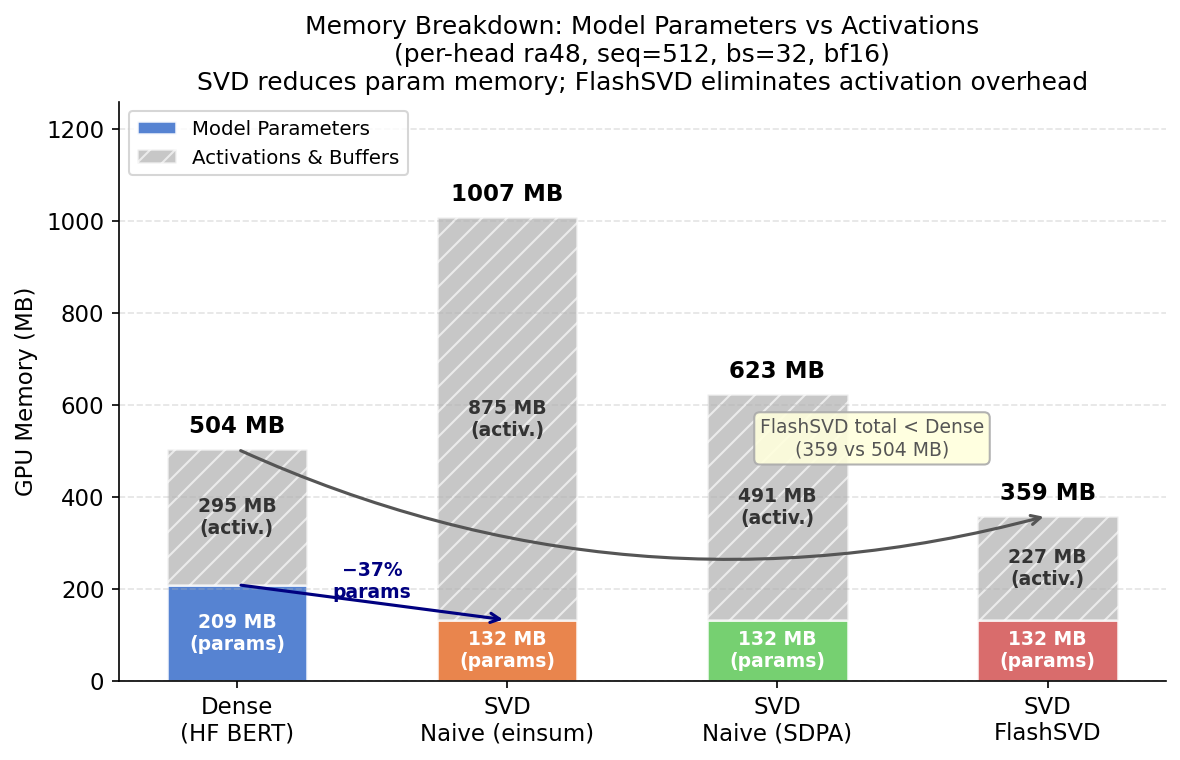}
        \caption{
        Memory breakdown into parameter storage and activation-related overhead for dense and compressed encoder inference.
        Results are shown at bf16, sequence length 512, and batch size 32.
        Compression reduces parameter memory, while fused backends further reduce total peak memory by lowering activation and buffer overhead.
        }
        \label{fig:memory_breakdown_appendix}
    \end{minipage}
\end{figure*}

Figure~\ref{fig:pareto_appendix} clarifies how backend effects should be interpreted relative to checkpoint quality. The vertical direction of the plot is determined by the compressed checkpoint, whereas the horizontal movement is largely determined by the runtime backend. In other words, the backend improves the efficiency of serving a fixed checkpoint without changing the checkpoint itself. This separation is useful conceptually and practically: compression methods compete on retained task quality, while backend design determines where that checkpoint sits on the efficiency axis. FlashSVD therefore complements, rather than replaces, checkpoint-level compression advances.

\begin{figure*}[!t]
    \centering
    \includegraphics[width=1\textwidth]{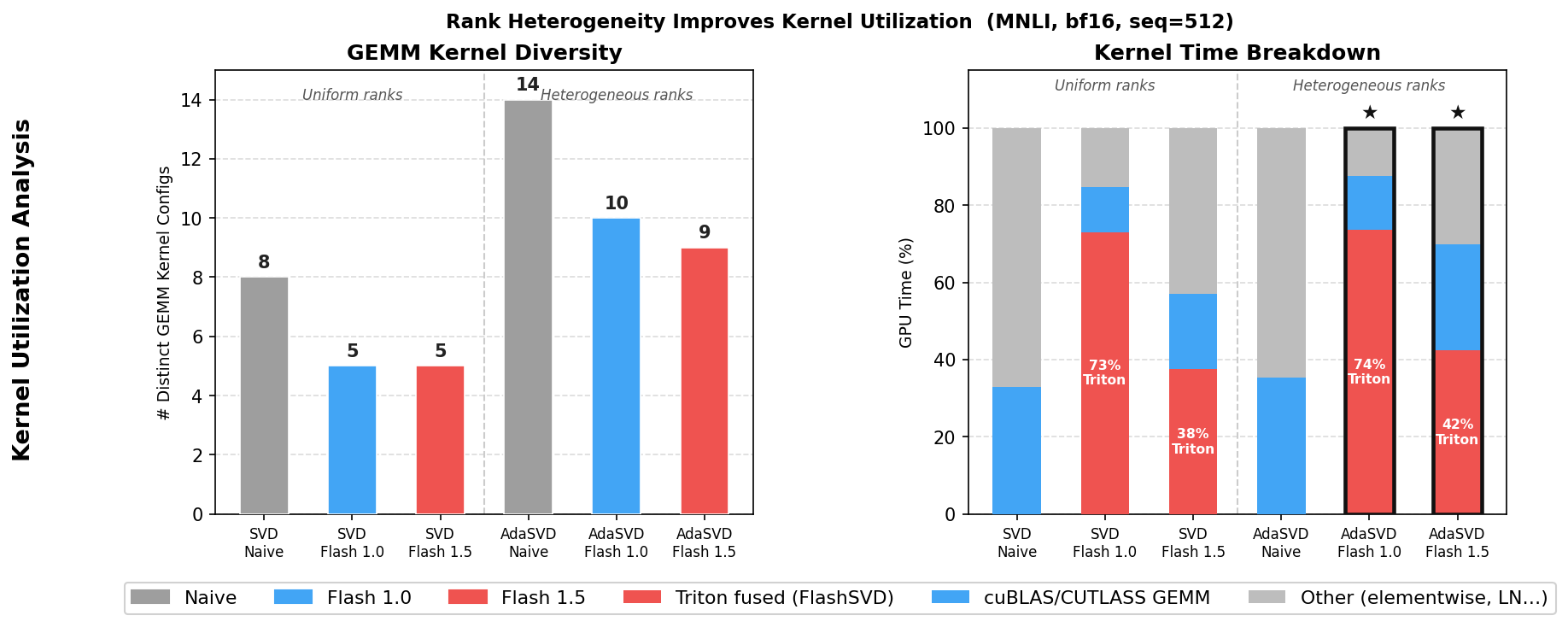}
    \caption{
    Kernel-level utilization analysis under uniform-rank and heterogeneous-rank compression on MNLI.
    The left panel reports the number of distinct GEMM kernel configurations, and the right panel shows GPU time breakdown across fused Triton kernels, cuBLAS/CUTLASS GEMM kernels, and other operations.
    We compare SVD and AdaSVD checkpoints under the naive backend, FlashSVD v1, and FlashSVD v1.5.
    Heterogeneous-rank compression increases kernel diversity under the naive backend, while both FlashSVD backends consolidate dispatch and shift execution toward fused kernels.
    }
    \label{fig:fig15_nsys_combined}
\end{figure*}

\FloatBarrier

\subsection{Mechanism Analysis}

Figure~\ref{fig:memory_breakdown_appendix} explains why compression alone is not enough to make encoder inference memory-efficient. Low-rank checkpoints reduce parameter storage, but naive low-rank execution can still incur large activation and rank-expansion overhead. The key difference is therefore not only what checkpoint is served, but how it is served. The FlashSVD backends lower total peak memory by eliminating much of the intermediate attention-related and rank-expansion buffering that remains present in the naive path. This is the encoder analogue of the decoder-side result that a better serving path, rather than low-rank arithmetic alone, determines practical efficiency.

Figure~\ref{fig:fig15_nsys_combined} shows the corresponding kernel-level mechanism. Under naive execution, encoder inference depends on a fragmented collection of generic GEMM calls, and heterogeneous-rank checkpoints increase this fragmentation further by introducing more kernel configurations. Both FlashSVD backends consolidate this structure into a smaller set of fused execution paths, reducing dispatch diversity and shifting more of the runtime into Triton-based fused kernels. This trend mirrors the decoder story: the speedup does not come only from reducing arithmetic, but from reorganizing the low-rank computation into a more stable and less fragmented runtime path.

\paragraph{Encoder summary.}
Taken together, the encoder results reinforce the same main lesson as the decoder results. FlashSVD backends improve practical efficiency not because low-rank checkpoints are automatically efficient, but because fused execution reduces activation overhead, consolidates fragmented dispatch, and scales more favorably as workloads grow. Among the tested backends, \texttt{flashsvd15} provides the strongest overall throughput/latency tradeoff, while the original \texttt{flashsvd} remains slightly more memory-minimal. The broader conclusion is the same across both model classes: practical low-rank acceleration is a runtime co-design problem as much as a compression problem.

\FloatBarrier

\section{Discussion}

In this work, we presented FlashSVD v1.5, demonstrating that realizing the theoretical benefits of low-rank compression requires eliminating the shattered serving path. By reorganizing fragmented execution into a streamlined dense-KV route and leveraging per-layer graph replay, we achieved significant inference acceleration for standard autoregressive LLMs and encoder models. 

While these results are strong, the underlying principle of execution defragmentation represents a generic serving paradigm that leaves substantial room for future exploration. A natural next step is extending this system-level co-design to Vision-Language Models (VLMs) and other multimodal inference settings~\cite{chen2025blind}. VLMs introduce additional systems complexities, such as cross-attention mechanisms and massive multimodal KV caches. Consolidating the fragmented execution paths of compressed vision encoders and cross-modal projectors will be critical to achieving true end-to-end acceleration in multimodal serving.

Similarly, generative diffusion models present a unique systems challenge. Whether utilizing traditional U-Net architectures or emerging Diffusion Transformers (DiTs), the generation process relies on heavily iterative denoising steps~\cite{jiangsada, wang2026accelerating, wang2026zeus}. Applying our low-rank runtime optimizations to diffusion models will require adapting offline prepacking and packed projections to manage their unique spatial feature maps, ensuring that kernel launch overhead does not bottleneck the multi-step generation process.

Finally, we plan to extend our support to non-Transformer architectures, most notably State-Space Models (SSMs) such as Mamba. Unlike standard Transformers, SSMs maintain implicit hidden states across time rather than an explicit token-by-token KV cache. Making low-rank SSMs fast will require designing entirely new memory layouts and operator fusions to prevent their recurrent state-update paths from becoming shattered. Ultimately, adapting our high-performance runtime to these diverse architectures will further establish that practical model compression must be fundamentally coupled with robust systems engineering and algorithmic optimization~\cite{cui2025efficient, shao2025scalable, wang2026enhanced}.


\end{document}